# Thinking Fast and Slow in Large Language Models

Thilo Hagendorff

thilo.hagendorff@iris.uni-stuttgart.de

University of Stuttgart

Sarah Fabi

sfabi@ucsd.edu

University of California San Diego

Michal Kosinski

michalk@stanford.edu

Stanford University

**Abstract** – Large language models (LLMs) are currently at the forefront of intertwining AI systems with human communication and everyday life. Therefore, it is of great importance to evaluate their emerging abilities. In this study, we show that LLMs like GPT-3 exhibit behavior that strikingly resembles human-like intuition – and the cognitive errors that come with it. However, LLMs with higher cognitive capabilities, in particular ChatGPT and GPT-4, learned to avoid succumbing to these errors and perform in a hyperrational manner. For our experiments, we probe LLMs with the Cognitive Reflection Test (CRT) as well as semantic illusions that were originally designed to investigate intuitive decision-making in humans. Our study demonstrates that investigating LLMs with methods from psychology has the potential to reveal otherwise unknown emergent traits.

**Introduction**

As the range of applications for large language models (LLMs) rapidly expands, it is of paramount importance to understand the mechanisms through which LLMs reason and make decisions. Recent research has revealed that with the increasing complexity of LLMs, they exhibit a multitude of emergent properties and abilities that were not explicitly anticipated or intended by their creators.[1] Among these newfound capabilities are the capacity to generate computer code, tackle mathematical problems, learn from examples, engage in introspection, solve theory of mind tasks, and a plethora of other skills.[2–6] In this paper, we aim to explore yet another emergent phenomenon, namely LLMs' reasoning capabilities, while shedding light on the intricacies of their cognitive processes.

Research on humans often distinguishes between two broad categories of reasoning or—more broadly—cognitive processes: Systems 1 and 2[7–14]. System 1 processes are fast, automatic, and instinctual. They often involve heuristics, or mental shortcuts, which enable quick judgments and decisions without conscious effort. System 1 is essential for everyday functioning, as it allows humans to navigate their environments and make rapid decisions with minimal effort. System 2 processes, on the other hand, are deliberate and require conscious effort. This system is employed in logical reasoning, critical thinking, and problem-solving. System 2 processes are slower and more resource-intensive, but they are also more accurate and less susceptible to bias.

On the surface, current-day LLMs are System 1 thinkers: The input text is processed by consecutive layers of neurons to produce a distribution of probabilities of all possible single-token (word) completions. This process is automatic, unidirectional, and involves a single wave of propagation through the neural network for each consecutive predicted word.[15] Current-day LLMs lack cognitive infrastructure (such as short-term memory), which is needed for re-examining the starting assumptions, testing alternative solutions, feedback loops, and other cognitive strategies typically associated with deliberate System 2 processes.[16] Yet, past research and the results presented in this paper suggest that, like humans, LLMs can engage in both System 1 and System 2 cognitive processes. While generating each consecutive word, LLMs re-read the *context*, including the original input (e.g., a task provided by a user) as well as the words they have thus far generated. As a result, the context may serve as an external short-term memory, which LLMs can use to engage in chain-of-thought reasoning, re-examine the starting assumptions, estimate partial solutions, or test alternative approaches. This is akin to how people use notepads to solve mathematical problems or write essays to sharpen and develop their arguments.

In this work, we build on psychological research on human reasoning and decision-making to explore System 1 and 2 processes in LLMs. We develop bespoke versions of two tests widely used in this field: the cognitive reflection test (CRT)[17] and semantic illusions[18] (see Table 1 and Table S1 in Supplementary Materials). The CRT comprises three types of mathematical tasks that appear to be simpler than they really are, thus triggering an intuitive but incorrect System 1 response. CRT Type 1 tasks, such as the widely known "A bat and a ball" task, use a "more than" phrase to trick participants into subtracting two of the values rather than solving a somewhat more complex equation. Type 2 tasks exploit people's tendency to complete a numerical triplet series, such as five machines making five widgets in five minutes because two machines make two widgets in two minutes. Type 3 tasks describe an exponential process but trick the participants into treating it as linear. Solving CRT tasks correctly requires engaging in deliberate System 2 reasoning or possessing well-developed System 1 intuitions. Semantic illusions are questions containing a disguised error aimed at triggering an intuitive but incorrect System 1 response. In the well-known Moses Illusion,[18] for example, participants tend to be tricked into claiming that Moses took two animals of each kind on the Ark (it was Noah).

Past research emphasized that LLMs tend to fall for the same traps as human participants in CRT tasks.[19] Here we show that LLMs can also engage in processes resembling deliberate System 2 reasoning. Moreover, we show that as LLMs' performance increases, their System-1-like next-word generation process becomes sufficiently acute to avoid falling for the traps embedded in the tasks. We also address some of the limitations of past studies. First, while past research focused on a single model (GPT-3), we study reasoning capabilities across a range of models of different sizes and complexities. Second, as solving CRT tasks requires mathematical abilities,[20] LLMs' performance could be limited by their mathematical skills. To address this issue, we complement the CRT tasks with semantic illusions that do not rely on mathematical skills. Third, past research relied on *three* CRT tasks copied verbatim from human studies[19]. This is problematic, as observing LLMs' performance on three CRT tasks does not allow for meaningful statistical comparisons. Moreover, these tasks (as well as their solutions) were likely present in LLMs' training data.[21] To circumvent these issues, we designed 50 bespoke versions of each type of task (200 in total; see Table S1).

For brevity and convenience, we use words such as "behavior," "intuition," "deliberation," or "ability" when referring to LLMs. Yet, we do not mean to equate artificial intelligence (AI) and human cognitive processes. While AI's outputs are often similar to the ones produced by humans, it typically operates in fundamentally different ways.

Table 1. The original tasks and examples of bespoke tasks written for our study.

| Original tasks | | | Our version (examples) | | |
|---|---|---|---|---|---|
| | Response | | | Response | |
| Prompt | Correct | Intuitive | Prompt | Correct | Intuitive |
| CRT Type 1 | | | | | |
| A bat and a ball cost $1.10 in total. The bat costs $1.00 more than the ball. How much does the ball cost? | $0.05 | $0.10 | A potato and a camera together cost $1.40. The potato costs $1 more than the camera. How much does the camera cost? | $0.20 | $0.40 |
| CRT Type 2 | | | | | |
| If it takes 5 machines 5 minutes to make 5 widgets, how long would it take 100 machines to make 100 widgets? | 5 minutes | 100 minutes | How long does it take 4 people to tailor 4 jackets, if it takes 7 people 7 hours to tailor 7 jackets? | 7 hours | 4 hours |

| CRT Type 3 | | | | | |
|---|---|---|---|---|---|
| In a lake, there is a patch of lily pads. Every day, the patch doubles in size. If it takes 48 days for the patch to cover the entire lake, how long would it take for the patch to cover half of the lake? | 47 days | 24 days | People are escaping from war. Each day, the total count of refugees doubles. If it takes 22 days for the entire population to evacuate, how long would it take for half of the population to do so? | 21 days | 11 days |
| Semantic Illusions | | | | | |
| How many animals of each kind did Moses take on the Ark? | It was Noah | Two | Where on their bodies do whales have their gills? | Whales do not have gills | On the sides of their heads |

**Methods**

Hypothesis-blind research assistants recruited on Upwork, a freelancing platform, prepared 50 semantic illusions and 50 CRT Type 3 tasks. CRT Type 1 and 2 tasks were generated automatically. All tasks can be found in Table S1. The code and data used here are available at https://osf.io/w5vhp/.

The tasks were administered to the family of OpenAI's Generative Pre-trained Transformer (GPT) models ranging from GPT-1 to ChatGPT-4.[6,22–24] To minimize the variance in the models' responses and thus increase the replicability of our results, the "temperature" parameter was set to 0. The response length was set to 100 tokens but was extended as needed. The models' responses were trimmed once they started repeating themselves or stopped responding to the task. The LLMs' responses were reviewed and scored manually. For clarity, the scoring is discussed in detail in the results section.

The same tasks were also administered to 500 human participants recruited on Prolific.io. Each participant was presented with a random set of four tasks (one of each kind) followed by a control question inquiring whether they used a language model or another external resource. (45 participants responded positively and were excluded from the analysis.) Human respondents' performance suggests that our tasks were of similar difficulty to those used in past human studies. In CRT, 38% of responses were correct, compared with 41% in the original study (n=3,428)[9] (difference ($\delta$) = 3%; $\chi^2 = 3.60; p = .06$). In semantic illusions, 64% of participants responded intuitively, compared with 52% in the original study (n=61; they did not report the fraction of correct responses; ($\delta = 12\%; \chi^2 = 2.41; p = .12$).[29]

**Study 1: Cognitive Reflection Test**

We first test the LLMs' performance on CRT tasks. To help the reader interpret the results, we discuss them in the context of LLMs' exemplary responses to CRT Type 3 task #14:

> *"In a cave, there is a colony of bats with a daily population doubling. Given that it takes 60 days for the entire cave to be filled with bats, how many days would it take for the cave to be half-filled with bats?"*

The *correct* response to this task is "59 days," but it was designed to appear easier than it really is, tempting participants to simply divide the total time by two, triggering an *intuitive* (but incorrect) response of "30 days." LLMs and humans' responses to this and other CRT tasks were divided into three categories: *correct*, *intuitive* (but incorrect), and *atypical*, which includes all other incorrect responses. Moreover, in each category of responses, the fraction of LLMs' responses preceded by a written chain-of-thought reasoning are marked by *black dots* (only one human described their reasoning process).

The overall performance of humans and LLMs across 150 CRT tasks is presented in Figure 1 (the Study 1 section). The results, split by CRT task types, are presented in Table S2. Four trends are apparent in Figure 1. First, most of the responses of early and smaller LLMs (up until GPT-3-curie) were classified as *atypical* (gray bars). This category includes responses that were evasive (e.g., GPT-1's response "*a lot*"); indicated failure to comprehend the task (e.g., GPT-2XL's response "*The colony would take 60 days to double in size*"); or were incorrect in ways different from one that the task was designed to trigger (e.g., GPT-3-babbage's response: "*It would take about 10 days for the cave to be half-filled with bats*"). Moreover, while 15% of responses of both GPT-3-babbage and GPT-3-curie were categorized as *correct* (green bars), they seemed accidental. All but one were given to CRT Type 2 tasks (see Table S2), which can be solved by simply repeating the number mentioned most frequently in the prompt—which these models tended to naively do in this and other tasks.

Second, as the models grew larger and their ability to comprehend the task increased, atypical responses were replaced by intuitive (but incorrect) responses (blue bars), which the tasks were designed to trigger (e.g., GPT-3-davinci-003: "*30 days*"). They constituted below 5% of responses of early models (up to GPT-3-babbage); increased to 21% for GPT-3-curie (difference ($\delta$) = 16%; $\chi^2 = 16.98; p < .001$); and further increased to 70%–90% for the GPT-3-davinci family ($\delta \geq 49\%; \chi^2 \geq 69.64; p < .001$), a fraction much higher than one observed in humans (55%; $\delta \geq 15\%; \chi^2 \geq 11.79; p < .001$).

In humans, such responses are interpreted as evidence of System 1 reasoning and failure to engage System 2, but they could also stem from deliberate—but erroneous—System 2 reasoning. The generative process behind the LLMs' responses is less ambiguous. As we discuss in the introduction, current-day LLMs lack the built-in cognitive infrastructure necessary to internally engage in System 2 processes. Consequently, their intuitive responses can only stem from a System-1-like process.

Importantly, LLMs' tendency to respond intuitively is unlikely to be driven by their insufficient mathematical skills. First, prior research showed that LLMs can solve basic mathematical problems.[2,5] Second, responding intuitively to CRT Type 1 and 3 tasks also requires solving a simple equation (e.g., how much is "half of 60"; responding intuitively to CRT Type 2 tasks requires no computation). Moreover, as we show in Study 3, GPT-3-davinci-003's performance can be significantly improved by presenting it with training examples.

Third, LLMs' strong tendency to respond intuitively stops abruptly with the arrival of ChatGPT. The fraction of correct responses equaled 59% for ChatGPT-3.5 and 96% for ChatGPT-4. This is much higher than 5% of tasks solved correctly by GPT-3-davinci-003, an otherwise very apt model ($\delta \geq 54\%; \chi^2 \geq 102.44; p < .001$) or 38% achieved by humans ($\delta \geq 21\%; \chi^2 \geq 25.60; p < .001$). ChatGPT's tendency to respond correctly was accompanied by a significant drop in their tendency to respond intuitively: 15% for ChatGPT-3.5 and 0% for ChatGPT-4 versus 80% for GPT-3-davinci-003 ($\delta \geq 65\%; \chi^2 \geq 125.81; p < .001$) and 55% for humans ($\delta \geq 40\%; \chi^2 \geq 86.30; p < .001$).

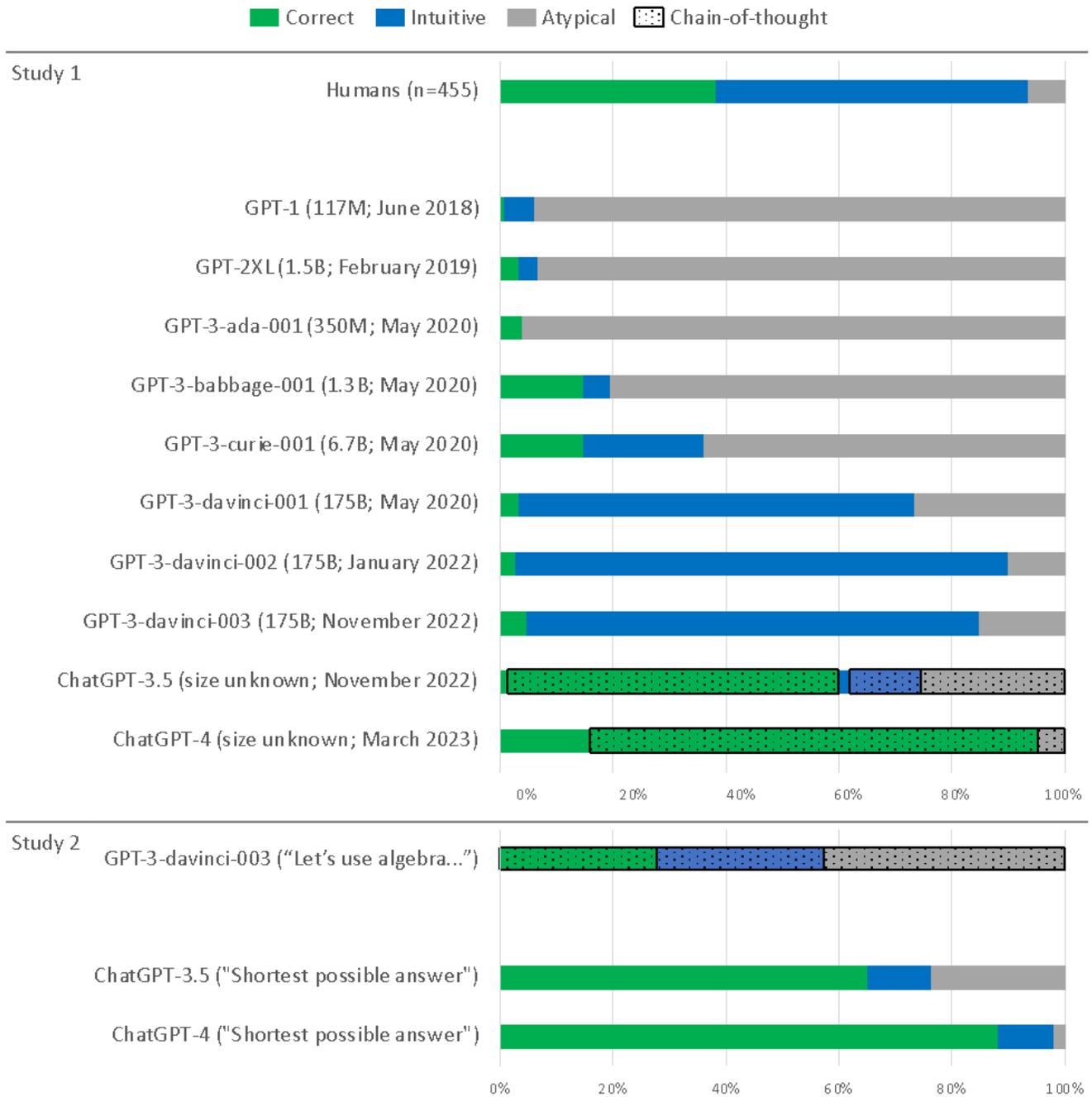

Figure 1. Overall human and LLM performance on CRT tasks.

A closer look at ChatGPT models' responses reveals that the dramatic increase in their performance was accompanied by a novel response style. While pre-ChatGPT LLMs responded with brief phrases or single sentences (e.g., GPT-3-davinci-003: "*30 days*"), 97% of ChatGPT-3.5's responses and 85% of ChatGPT-4.5's responses included some form of chain-of-thought reasoning (such responses are marked with black dots on bars on Figure 1). See, for example, ChatGPT-3.5's response to CRT Type 1 task #14:

> ChatGPT-3.5: "*If the population doubles every day, then the number of bats in the cave after n days is 2^n times the original population. Let P be the original population of the colony.*
> *After 60 days, the population is 2^60 \* P, which is the capacity of the cave.*
> *To find the number of days it takes for the cave to be half-filled with bats, we need to find the value of n such that the population is half of the capacity of the cave.*
> *0.5 \* 2^60 \* P = 2^n \* P*
> *Dividing both sides by P, we get:*
> *0.5 \* 2^60 = 2^n*
> *Taking the logarithm base 2 of both sides, we get:*
> *n = 60 + log2(0.5) = 60 - 1 = 59*
> *Therefore, it would take 59 days for the cave to be half-filled with bats.*"

As we discussed before, there is nothing deliberate in how ChatGPT-3.5 and other current-day LLMs generate the next word. Yet, each time the word is generated, LLM re-reads the task and the response it generated so far, combining System-1-like next-word generation into something *resembling* a System 2 process: generating a strategy needed to solve the task, dividing the task into more tractable subtasks, and solving them one by one. This is akin to how humans use a notebook to solve mathematical tasks without the need to process them in their short-term memory. In Study 2, we will show evidence suggesting that chain-of-thought responses not only resemble but also serve as a System 2 process.

**Study 2: Chain-of-Thought Reasoning in CRT Tasks**
The results of Study 1 show that the abrupt increase in LLMs' performance on CRT tasks was accompanied by their employment of the chain-of-thought reasoning. Past research suggests that instructing LLMs to engage in chain-of-thought reasoning improves their ability to solve various tasks.[3] Could it be that we observed a similar phenomenon here?

To answer this question, we first instruct GPT-3-davinci-003 to engage in the chain-of-thought reasoning by adding the following suffix after the prompt: "Let's use algebra to solve this problem." The results presented in Figure 1 (in the Study 2 section) show that our manipulation was successful: The fraction of chain-of-thought responses increased from 0% in Study 1 to 100% ($\delta = 100\%; \chi^2 = 147.01; p < .001$). The model seemed to design and execute a task-solving strategy. In some cases, it was sound, boosting the fraction of correct responses from 5% to 28% ($\delta = 23\%; \chi^2 = 28.20; p < .001$). Most of the time, it was poorly conceived or executed, leading to the increase of atypical responses from 15% to 43% ($\delta = 28\%; \chi^2 = 14.72; p < .001$). Most importantly, it significantly reduced a model's tendency to fall for the trap embedded in the task, as illustrated by the drop in the fraction of intuitive responses from 80% to 29% ($\delta = 51\%; \chi^2 = 75.66; p < .001$).

Next, we prevent ChatGPT models from engaging in chain-of-thought reasoning by adding the following suffix: "Provide the shortest possible answer (e.g., '$2' or '1 week'), do not explain your

reasoning." The results presented in Figure 1 (the Study 2 section) show that our manipulation was successful: The fraction of chain-of-thought responses fell from 97% to 0% for ChatGPT-3.5 ($\delta = 97\%; \chi^2 = 276.79; p < .001$) and from 84% to 0% for ChatGPT-4 ($\delta = 84\%; \chi^2 = 213.81; p < .001$). The fraction of correct responses for ChatGPT-3.5 did not significantly change ($\delta = 4\%; \chi^2 = 0.47; p = .49$). For ChatGPT-4, the fraction of intuitive responses increased from 0% to 10% ($\delta = 10\%; \chi^2 = 13.75; p < .001$), and correct responses fell from 95% to 88% ($\delta = 7\%; \chi^2 = 4.36; p < .05$).

The results of Study 2 suggest that chain-of-thought reasoning helps LLMs to avoid falling for the traps embedded in the CRT tasks and improves their ability to solve them correctly. Yet, they also reveal that ChatGPT models can solve a great majority of CRT tasks even when forced to provide a System-1-like response. This suggests that ChatGPT models have well-developed intuition, enabling them to solve CRT tasks without engaging in chain-of-thought reasoning. This is consistent with ChatGPT-4's performance in Study 1, where it solved 24% of the CRT task without using chain-of-thought reasoning. Study 3 explores this phenomenon in greater detail.

**Study 3: Improving LLMs' Intuition in CRT Tasks**

The results of Study 2 suggest that the most recent of the LLMs examined here—the ChatGPT family—can avoid falling for the trap embedded in CRT tasks and solve them correctly, even when prevented from engaging in the chain-of-thought System-2-like reasoning. In humans, this would be taken as evidence of a well-developed intuition stemming from previous exposure to CRT or similar tasks[25,26] (although the persistence and size of this effect is disputed[27,28]). Here we show results suggesting that the same applies to LLMs. This is in line with past results showing that LLMs can learn, even from a single example.

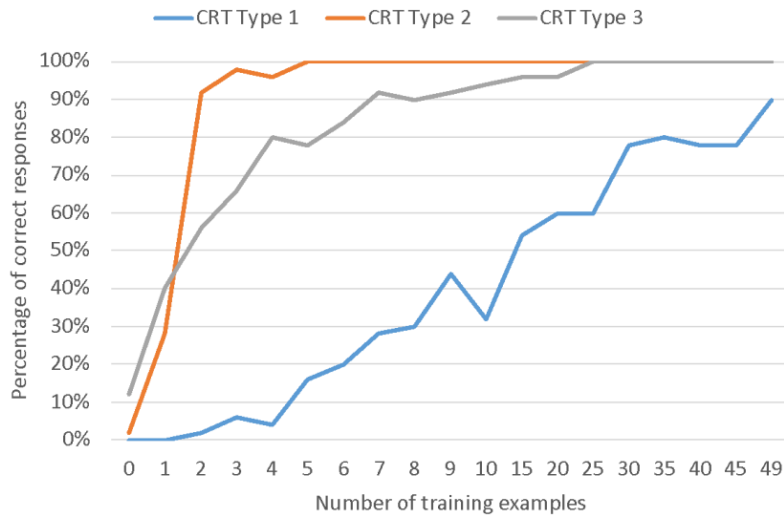

Figure 2. Change in the fraction of correct responses of GPT-3-davinci-003 against the number of training examples that the task was prefixed with.

As ChatGPT models already seem to possess well-developed intuition, we attempt to improve the System-1-like responses of GPT-3-davinci-003. We present it with CRT tasks, each time preceding it with 0 to 49 remaining tasks of the same type, accompanied by the correct solution. CRT tasks of the same type are semantically very similar, enabling the model to develop System-1 intuitions akin to one expressed by the ChatGPT model family.

The results presented in Figure 2 show that GPT-3-davinci-003's ability to answer correctly (rather than intuitively) increased with each additional example. The fastest gains were observed for CRT Type 2 tasks, where the accuracy increases from 2% to 92% after two examples ($\delta = 90\%; \chi^2 = 77.72; p < .001$). This is to be expected, as they can be solved correctly by simply repeating the duration listed in the task. To solve CRT Type 3, one needs to respond with the total time minus one unit. They proved to be somewhat more complex: The accuracy increased from 12% to 92% after seven training examples ($\delta = 80\%; \chi^2 = 60.94; p < .001$). It took most examples to develop the model's intuition to solve CRT Type 1 tasks, where the correct answer equals $\frac{total\ price - known\ price}{2}$. But even here, the model's accuracy increased from 0% to about 78% accuracy after 30 examples ($\delta = 78\%; \chi^2 = 60.70; p < .001$).

**Study 4: Semantic Illusions**

CRT tasks employed in Studies 1 to 3 rely heavily on mathematical skills and are highly semantically uniform. To make sure that the results generalize beyond the CRT tasks, we examine LLMs' performance on much more semantically diverse tasks: semantic illusions. Like CRT tasks, semantic illusions contain a disguised error aimed at triggering an intuitive but incorrect System 1 response. Unlike CRT tasks, semantic illusions do not require mathematical skills, instead relying on participants' general knowledge.

To help the reader interpret the results, we discuss them in the context of LLMs' exemplary responses to semantic illusion #47:

> "*Which famous artist designed the famous church, la Sagrada Familia, located in Madrid?*"

As in the context of CRT tasks, responses were divided into three categories: *intuitive*, *correct*, and *atypical*. The question was designed to trigger an *intuitive* System-1 response "*Antoni Gaudí*" while overlooking the embedded invalid assumption (e.g., la Sagrada Familia is in Barcelona). Yet, responding "*Antoni Gaudí*" can be treated as indicative of System 1 processing only if the respondent has the knowledge necessary to recognize the error. Thus, given an intuitive response, the model was reset, and its underlying knowledge was tested using an additional question (here: "*Where is the famous church, la Sagrada Familia, located?*"; see Table S3). Intuitive responses given by LLMs that failed this post-hoc test were recategorized as *atypical*, along with responses revealing a further lack of necessary knowledge (e.g., GPT-3-babbage: "*Francisco Goya*") and nonsensical responses (e.g., GPT-1 "*the church of san francisco*"). Responses that recognized that this question is invalid were categorized as *correct* (e.g., ChatGPT-4's response: "*La Sagrada Familia is actually located in Barcelona, not Madrid, and was designed by the famous Spanish architect Antoni Gaudí.*").

The results presented in Figure 3 show a pattern similar to the one observed in the context of CRT tasks. Most of the responses of early and smaller LLMs (up to GPT-3-babbage) were atypical (gray bars), as they struggled to comprehend the question or lacked the necessary knowledge. As LLMs grow in size and sophistication, the fraction of atypical responses fell from 52% for GPT-3-babbage to 10% for GPT-3-davinci-003 ($\delta = 42\%$; $\chi^2 = 18.70$; $p < .001$). They were replaced by intuitive responses (blue bars): GPT-3-davinci-003 fell for the semantic illusion 72% of the time. As in the context of CRT tasks, this trend changes dramatically with the introduction of ChatGPT. The fraction of correct responses increased from 18% for GPT-3-davinci-003 to 74% and 88% for ChatGPT-3.5 and ChatGPT-4, respectively (green bars; $\delta \geq 56\%$; $\chi^2 = 29.35$; $p < .001$). As we discussed before, there is nothing deliberate in LLMs' next-word generation process. Yet, this System-1-like process proved to be very apt at detecting invalid assumptions embedded in semantic illusions.

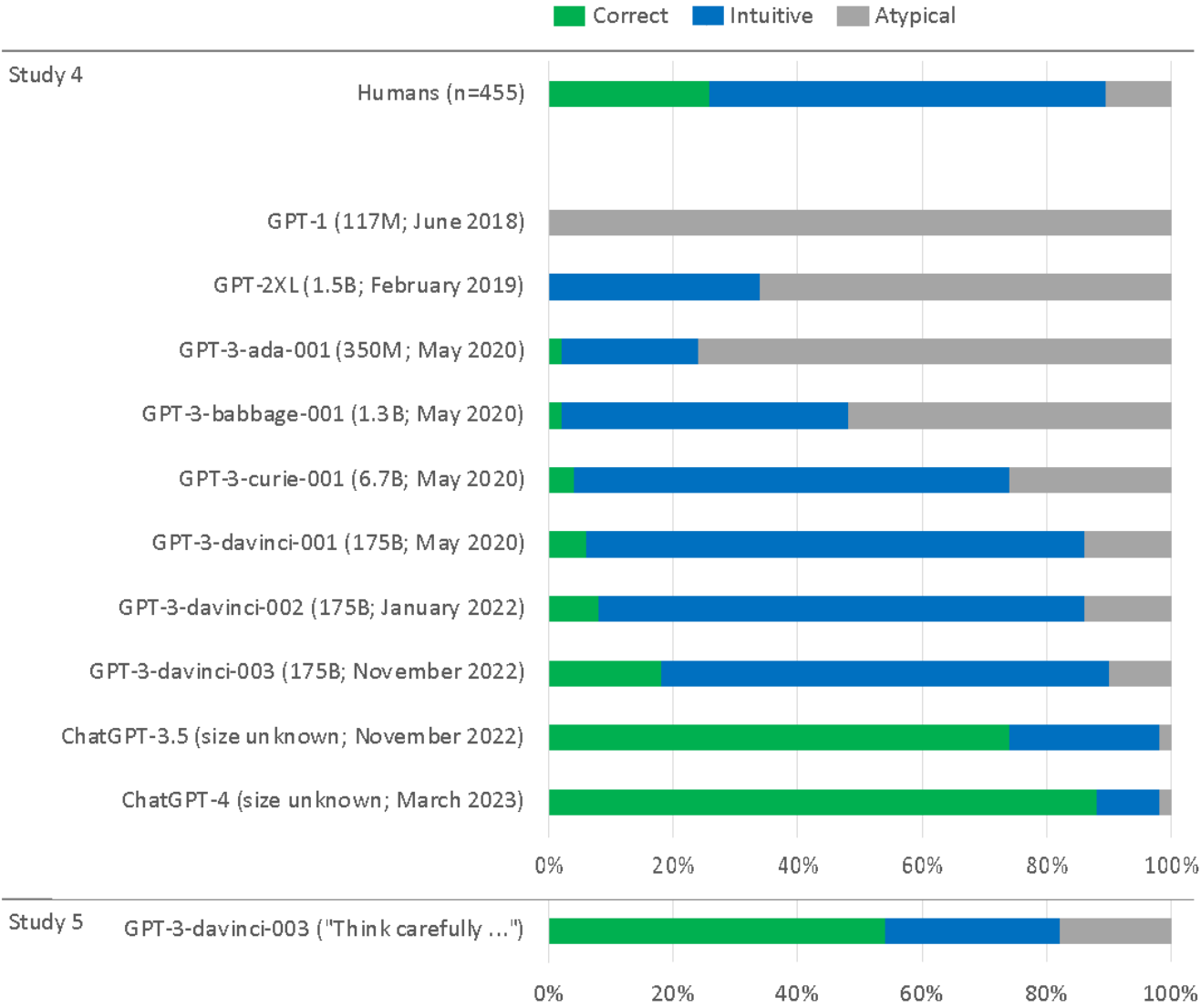

Figure 3. Overall human and LLM performance in semantic illusions tasks. Confidence intervals are reported in **Table S1**.

**Study 5: Improving LLMs' Performance on Semantic Illusions**
The results of Studies 2 and 3 suggest that LLMs' propensity to commit reasoning errors in CRT tasks can be reduced by instructing them to examine the task more carefully and providing them with examples of correct solutions to similar tasks. Study 5 replicates those results in the context of semantic illusions.

We first add the suffix "*Think carefully and check the question for invalid assumptions*" to each semantic illusion and administer them to GPT-3-davinci-003. The results presented in Figure 3 (the Study 5 section) show that the fraction of correct responses increased threefold, from 18% in Study 4 to 54% ($\delta = 36\%; \chi^2 = 12.54; p < .001$), while the fraction of intuitive responses decreased from 72% to 28% ($\delta = 44\%; \chi^2 = 17.64; p < .001$).

Next, as in Study 3, we precede each semantic illusion with 0 to 49 other semantic illusions, accompanied by the correct solution. The results presented in Figure 4 show that GPT-3-davinci-003's ability to answer correctly increased from 18% for 0 examples to over 64% for 10 and more examples ($\delta \geq 46\%; \chi^2 = 20.01; p < .001$).

In summary, LLMs' performance on semantic illusions mirrored the one observed in CRT: The early models struggled with comprehending the tasks, the later models showed an above-human tendency to respond intuitively (but incorrectly), and the most recent ChatGPT models showed an above-human ability to avoid succumbing to semantic illusions. Moreover, as in the context of CRT tasks, GPT-3-davinci-003's performance could be significantly improved by instructing it to engage in careful reasoning and by boosting its intuition by exposing it to training examples.

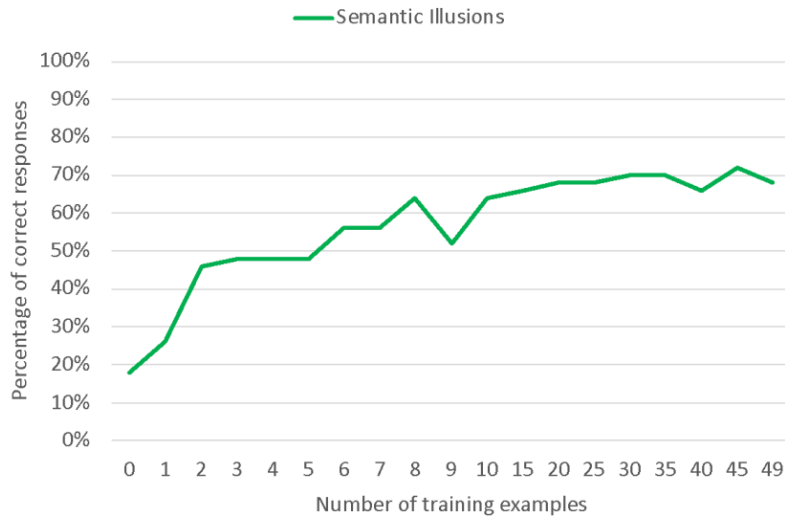

Figure 4. Change in the fraction of correct responses of GPT-3-davinci-003 against the number of training examples that the task was prefixed with.

**Discussion**

In this work, we studied reasoning processes in LLMs using CRT and semantic illusions, widely used to study reasoning processes in humans. Our results reveal three major patterns. First, Studies 1 and 4 show that as LLMs grow in size and their ability to comprehend the tasks increase, they tend to fall for the trap embedded in the tasks and respond intuitively—which, in humans, would be interpreted as evidence of fast, automatic, and instinctual System 1 processing. GPT-3-davinci-003, for example, responded intuitively in 80% CRT tasks and 72% semantic illusions, a fraction higher than the one observed in humans (55% and 64%, respectively). How would one explain pre-ChatGPT models' tendency to respond intuitively, despite their sufficient mathematical abilities and factual knowledge demonstrated in Studies 3, 4, and 5? As we discuss in the introduction, LLMs lack the cognitive infrastructure necessary to engage in System 2 processes, which some humans may successfully activate when answering such questions. Thus, in the absence of well-developed intuition or explicit chain-of-thought reasoning, they are particularly prone to fall for the traps embedded in the tasks.

Second, the tendency to respond intuitively changed abruptly with the arrival of ChatGPT models, which responded correctly to a great majority of tasks, avoiding semantic traps embedded in them. ChatGPT-4, for example, responded correctly to 95% of CRT tasks and 88% of semantic illusions, compared with 38% and 26% in humans, respectively. The majority of ChatGPT's responses to CRT tasks involved chain-of-thought reasoning, where the models developed strategies needed to solve the task, examined the starting assumptions, estimated partial solutions, or tested alternative approaches. This confirms past results showing that LLMs can engage in System-2-like reasoning using their input-output context window, in a way akin to how people use notepads to solve mathematical problems or write essays to develop their arguments. The results of Study 2 further confirm that chain-of-thought reasoning can boost LLMs' performance: The fraction of GPT-3-davinci-003's correct responses to CRT tasks increased from 5% to 28% when it was made to engage in chain-of-thought reasoning in its output-input context window.

Third, further analysis reveals that while ChatGPT models tended to engage in chain-of-thought reasoning in their responses to CRT tasks, they could solve both CRT tasks and semantic illusions without it. Study 2 shows that ChatGPT models answer correctly to the great majority of CRT tasks even when prevented from engaging in chain-of-thought reasoning. Also, Study 4 shows that their responses to semantic illusions were both predominantly correct and did not include chain-of-thought reasoning. This suggests that ChatGPT models have well-developed intuitions enabling them to solve CRT tasks and semantic illusions without engaging System-2 like processes. This finding is further supported by Studies 3 and 5 showing that GPT-3-davinci-003's performance on both CRT tasks and semantic illusions can be significantly increased by presenting it with example tasks and their correct solutions.

What drove the steep shift in accuracy and response style between GPT-3 and ChatGPT? Some progress is to be expected. In humans, CRT and semantic illusions are good predictors of ability to engage in unbiased, reflective, and rational decision-making,[29–32] as well as overall cognitive ability.[17,32] Thus, LLMs' ability to solve CRT and semantic illusions should increase as their overall ability increases. Yet, the shift observed in this study seems to be steeper than the increase in LLMs' overall abilities. We can only speculate, given that OpenAI shares limited information on the technical specification and training process of their models. First, it is unlikely that the shift was driven merely by larger model size. According to OpenAI, ChatGPT-3.5-turbo was derived from text-davinci-003 by fine-tuning it for chat. Both models are likely of similar size. Second, it could be that the shift was driven by the employment of reinforcement learning from human feedback (RLHF).[33,34] In RLHF, human-written demonstrations on example prompts are used to train supervised learning baselines. Next, human "AI trainers" rank model outputs on a larger set of API prompts, and a reward model is trained to predict their preferences. This reward model is then used to fine-tune the models using Proximal Policy Optimization Algorithms.[35] While RLHF was employed since GPT-3 text-davinci-002,[33] this procedure was enhanced in ChatGPT training: AI trainers played both sides—the user and an AI assistant.[36] Third, it is likely that ChatGPT models were exposed to sufficient CRT-like tasks in their training to be able to respond to them intuitively. Those tasks are highly semantically similar and, as the results of Study 3 illustrate, exposure to training examples can rapidly boost an LLM's ability to solve them correctly. This explanation is less likely in the context of semantic illusions, which are much more irregular and diverse. This question will hopefully be addressed by further research or more transparency in LLMs' development.

Next to the analysis of LLMs' performance on reasoning tasks, one can approach the issue from a normative perspective, asking whether phenomena of intuitive decision-making are desirable in LLMs. In the cognitive science literature, researchers stress that the notion of intuitive errors relies on a normative concept of logics and statistics, which can be inappropriate for real-world situations. Instead, decision-making processes should be evaluated in the sense of "ecological rationality," meaning based on how well they fit the structure of the environment in which they occur.[37–39] In this vein, CRT tasks as well as semantic illusions create a "hostile" test environment, which intentionally aims to mislead humans. Regarding an ecologically rational response to semantic illusions, one can ask: Should LLMs go with the conversational flow and just "overlook" small mistakes instead of correcting factually incorrect questions? In conversations between humans, rapid turn-taking between communication partners demonstrates high degrees of communicative coordination, which avoids interruption.[40] This process is likely to fail if semantic structures would be continuously checked in a verbatim analysis. If language models behave similarly, this would mean that they promote fluent communication, which might be an important capability as soon as these models are natural interlocutors in ordinary social contexts. On the other hand, "hallucinated" outputs, factual incorrectness, and misinformation harms are a major problem in LLMs.[41,42] Nevertheless, and apart from normative discussions regarding the evaluation and intuitive "errors," it remains noteworthy that even in hostile test environments such as the CRT or semantic illusions, state-of-the-art LLMs such as ChatGPT and GPT-4 can (in most cases) be triggered to not make mistakes. This speaks for their remarkable reasoning abilities, no matter whether these reflect "logical intuitions"[7] or deliberate reasoning, which are likely to have effects in numerous areas other than our test environment.

The progress in LLMs not only increased their capabilities, but also reduced our ability to anticipate their properties and behavior. It is increasingly difficult to study LLMs through the lenses of their architecture and hyperparameters.[19,43,44] Instead, as we show in this work, LLMs can be studied using methods designed to investigate another capable and opaque structure, namely the human mind.[45,46] Our approach falls within a quickly growing category of studies employing classical psychological tests and experiments to probe LLMs' "psychological" processes such as judgment, decision-making, or cognitive biases.[19,44,47–49]

**Code availability**

The code as well as all datasets are publicly available under https://osf.io/w5vhp/.

**Author Contributions**:

TH, SF, and MK wrote the manuscript, conducted the experiments, and interpreted the results. TH had the idea for the paper.

**Supplementary Materials**

Table S1. List of all CRT tasks and semantic illusions.

| CRT Type 1 Tasks |
|---|
| Task #1: A pear and a fridge together cost $140. The pear costs $100 more than the fridge. How much does the fridge cost?<br>Task #2: A potato and a camera together cost $1.40. The potato costs $1 more than the camera. How much does the camera cost?<br>Task #3: A boat and a potato together cost $110. The boat costs $100 more than the potato. How much does the potato cost?<br>Task #4: A light bulb and a pan together cost $12. The light bulb costs $10 more than the pan. How much does the pan cost?<br>Task #5: A chair and a coat together cost $13. The chair costs $10 more than the coat. How much does the coat cost?<br>Task #6: A tube of toothpaste and a wallet together cost $54. The tube of toothpaste costs $50 more than the wallet. How much does the wallet cost? |

Task #7: A coffee table and a mixing bowl together cost $52. The coffee table costs $50 more than the mixing bowl. How much does the mixing bowl cost?
Task #8: A microwave oven and a flower box together cost $51. The microwave oven costs $50 more than the flower box. How much does the flower box cost?
Task #9: A can of baby food and a helmet together cost $1.20. The can of baby food costs $1 more than the helmet. How much does the helmet cost?
Task #10: A sculpture and a box of lipstick together cost $13. The sculpture costs $10 more than the box of lipstick. How much does the box of lipstick cost?
Task #11: A pair of leggings and a body lotion together cost $54. The pair of leggings costs $50 more than the body lotion. How much does the body lotion cost?
Task #12: A trampoline and a box of batteries together cost $520. The trampoline costs $500 more than the box of batteries. How much does the box of batteries cost?
Task #13: A mouse and a can of baby food together cost $1.20. The mouse costs $1 more than the can of baby food. How much does the can of baby food cost?
Task #14: A bottle of shampoo and a knife together cost $110. The bottle of shampoo costs $100 more than the knife. How much does the knife cost?
Task #15: A bottle of bath salts and a set of crayons together cost $520. The bottle of bath salts costs $500 more than the set of crayons. How much does the set of crayons cost?
Task #16: A set of wine glasses and a pencil together cost $12. The set of wine glasses costs $10 more than the pencil. How much does the pencil cost?
Task #17: A wallet and a camera together cost $52. The wallet costs $50 more than the camera. How much does the camera cost?
Task #18: A pair of hiking boots and a coffee table together cost $1.20. The pair of hiking boots costs $1 more than the coffee table. How much does the coffee table cost?
Task #19: A tube of toothpaste and a pear together cost $12. The tube of toothpaste costs $10 more than the pear. How much does the pear cost?
Task #20: A trampoline and a flower box together cost $54. The trampoline costs $50 more than the flower box. How much does the flower box cost?
Task #21: A silicone case and a lawnmower together cost $130. The silicone case costs $100 more than the lawnmower. How much does the lawnmower cost?
Task #22: A toy and a microwave oven together cost $5.40. The toy costs $5 more than the microwave oven. How much does the microwave oven cost?
Task #23: A purse and a mixing bowl together cost $5.40. The purse costs $5 more than the mixing bowl. How much does the mixing bowl cost?
Task #24: A hat and a mouse together cost $540. The hat costs $500 more than the mouse. How much does the mouse cost?
Task #25: A purse and a box of cigarettes together cost $1.40. The purse costs $1 more than the box of cigarettes. How much does the box of cigarettes cost?
Task #26: A box of mascara and a microwave oven together cost $14. The box of mascara costs $10 more than the microwave oven. How much does the microwave oven cost?
Task #27: A bag and a food processor together cost $54. The bag costs $50 more than the food processor. How much does the food processor cost?
Task #28: A coat and a pot together cost $510. The coat costs $500 more than the pot. How much does the pot cost?

Task #29: A jacket and a set of wine glasses together cost $5.10. The jacket costs $5 more than the set of wine glasses. How much does the set of wine glasses cost?
Task #30: A hat and a coffee table together cost $1.20. The hat costs $1 more than the coffee table. How much does the coffee table cost?
Task #31: A pencil and a keyboard together cost $14. The pencil costs $10 more than the keyboard. How much does the keyboard cost?
Task #32: A bag and a coffee table together cost $53. The bag costs $50 more than the coffee table. How much does the coffee table cost?
Task #33: A mixing bowl and a laundry detergent together cost $52. The mixing bowl costs $50 more than the laundry detergent. How much does the laundry detergent cost?
Task #34: A perfume and a bottle of ouzo together cost $120. The perfume costs $100 more than the bottle of ouzo. How much does the bottle of ouzo cost?
Task #35: A bottle of bath salts and a pack of diapers together cost $510. The bottle of bath salts costs $500 more than the pack of diapers. How much does the pack of diapers cost?
Task #36: A CD and a pan together cost $13. The CD costs $10 more than the pan. How much does the pan cost?
Task #37: A rice cooker and a sculpture together cost $130. The rice cooker costs $100 more than the sculpture. How much does the sculpture cost?
Task #38: A pair of sunglasses and a toy together cost $13. The pair of sunglasses costs $10 more than the toy. How much does the toy cost?
Task #39: A speaker and a bottle of ouzo together cost $5.20. The speaker costs $5 more than the bottle of ouzo. How much does the bottle of ouzo cost?
Task #40: A set of crayons and a wallet together cost $5.20. The set of crayons costs $5 more than the wallet. How much does the wallet cost?
Task #41: A rug and a sculpture together cost $54. The rug costs $50 more than the sculpture. How much does the sculpture cost?
Task #42: A pair of sunglasses and a light bulb together cost $130. The pair of sunglasses costs $100 more than the light bulb. How much does the light bulb cost?
Task #43: A scarf and a ring together cost $1.20. The scarf costs $1 more than the ring. How much does the ring cost?
Task #44: A pair of leggings and a scarf together cost $52. The pair of leggings costs $50 more than the scarf. How much does the scarf cost?
Task #45: A wall clock and a light bulb together cost $1.30. The wall clock costs $1 more than the light bulb. How much does the light bulb cost?
Task #46: A blanket and a flashlight together cost $52. The blanket costs $50 more than the flashlight. How much does the flashlight cost?
Task #47: A keyboard and a pair of socks together cost $54. The keyboard costs $50 more than the pair of socks. How much does the pair of socks cost?
Task #48: A set of wine glasses and a pizza together cost $53. The set of wine glasses costs $50 more than the pizza. How much does the pizza cost?
Task #49: A pear and a bottle of ouzo together cost $54. The pear costs $50 more than the bottle of ouzo. How much does the bottle of ouzo cost?
Task #50: A pizza and a toy together cost $13. The pizza costs $10 more than the toy. How much does the toy cost?

| CRT Type 2 Tasks |
|---|
| Task #1: How long does it take 4 people to tailor 4 jackets, if it takes 7 people 7 hours to tailor 7 jackets? |
| Task #2: How long does it take 4 washing machines to wash 4 loads of laundry, if it takes 8 washing machines 8 hours to wash 8 loads of laundry? |
| Task #3: How long does it take 50 bees to pollinate 50 flowers, if it takes 60 bees 60 minutes to pollinate 60 flowers? |
| Task #4: How long does it take 1 carpenter to make 1 chair, if it takes 5 carpenters 5 days to make 5 chairs? |
| Task #5: How long does it take 10 ovens to bake 10 lasagnas, if it takes 60 ovens 60 minutes to bake 60 lasagnas? |
| Task #6: How long does it take 1 researcher to publish 1 paper, if it takes 6 researchers 6 years to publish 6 papers? |
| Task #7: How long does it take 30 cleaners to clean 30 rooms, if it takes 50 cleaners 50 hours to clean 50 rooms? |
| Task #8: How long does it take 40 students to change 40 light bulbs, if it takes 70 students 70 minutes to change 70 light bulbs? |
| Task #9: How long does it take 40 builders to build 40 houses, if it takes 50 builders 50 weeks to build 50 houses? |
| Task #10: How long does it take 5 people to plant 5 trees, if it takes 6 people 6 minutes to plant 6 trees? |
| Task #11: How long does it take 30 coffee machines to make 30 coffees, if it takes 40 coffee machines 40 minutes to make 40 coffees? |
| Task #12: How long does it take 5 machines to pack 5 boxes of chocolates, if it takes 8 machines 8 minutes to pack 8 boxes of chocolates? |
| Task #13: How long does it take 10 children to eat 10 boxes of chocolates, if it takes 50 children 50 minutes to eat 50 boxes of chocolates? |
| Task #14: How long does it take 2 people to read 2 books, if it takes 4 people 4 weeks to read 4 books? |
| Task #15: How long does it take 5 teams to renovate 5 houses, if it takes 8 teams 8 weeks to renovate 8 houses? |
| Task #16: How long does it take 30 people to knit 30 pairs of socks, if it takes 40 people 40 weeks to knit 40 pairs of socks? |
| Task #17: How long does it take 40 people to pick 40 fields of strawberries, if it takes 70 people 70 hours to pick 70 fields of strawberries? |
| Task #18: How long does it take 6 programmers to write 6 lines of code, if it takes 7 programmers 7 hours to write 7 lines of code? |
| Task #19: How long does it take 4 photographers to take 4 photos, if it takes 8 photographers 8 hours to take 8 photos? |
| Task #20: How long does it take 1 painter to paint 1 painting, if it takes 8 painters 8 hours to paint 8 paintings? |
| Task #21: How long does it take 50 writers to write 50 books, if it takes 70 writers 70 minutes to write 70 books? |

Task #22: How long does it take 2 cooks to cook 2 meals, if it takes 8 cooks 8 minutes to cook 8 meals?
Task #23: How long does it take 50 doctors to examine 50 patients, if it takes 60 doctors 60 minutes to examine 60 patients?
Task #24: How long does it take 2 drivers to change 2 tires, if it takes 7 drivers 7 minutes to change 7 tires?
Task #25: How long does it take 2 farm workers to pick 2 apples, if it takes 8 farm workers 8 seconds to pick 8 apples?
Task #26: How long does it take 4 freezers to freeze 4 liters of water, if it takes 6 freezers 6 hours to freeze 6 liters of water?
Task #27: How long does it take 20 bakers to bake 20 cakes, if it takes 80 bakers 80 hours to bake 80 cakes?
Task #28: How long does it take 30 hair stylists to finish 30 hairstyles, if it takes 50 hair stylists 50 minutes to finish 50 hairstyles?
Task #29: How long does it take 2 mechanics to fix 2 cars, if it takes 3 mechanics 3 hours to fix 3 cars?
Task #30: How long does it take 20 tailors to make 20 dresses, if it takes 50 tailors 50 hours to make 50 dresses?
Task #31: How long does it take 3 painters to paint 3 rooms, if it takes 4 painters 4 hours to paint 4 rooms?
Task #32: How long does it take 40 trees to grow 40 leaves, if it takes 60 trees 60 days to grow 60 leaves?
Task #33: How long does it take 1 runner to clean 1 shoe, if it takes 5 runners 5 minutes to clean 5 shoes?
Task #34: How long does it take 3 translators to translate 3 pages, if it takes 6 translators 6 hours to translate 6 pages?
Task #35: How long does it take 2 machines to make 2 smartphones, if it takes 3 machines 3 hours to make 3 smartphones?
Task #36: How long does it take 50 people to smoke 50 cigarettes, if it takes 70 people 70 minutes to smoke 70 cigarettes?
Task #37: How long does it take 7 opticians to make 7 glasses, if it takes 8 opticians 8 days to make 8 glasses?
Task #38: How long does it take 40 pipes to fill 40 containers, if it takes 60 pipes 60 hours to fill 60 containers?
Task #39: How long does it take 50 people to eat 50 pizzas, if it takes 80 people 80 minutes to eat 80 pizzas?
Task #40: How long does it take 20 kettles to boil 20 liters of water, if it takes 80 kettles 80 hours to boil 80 liters of water?
Task #41: How long does it take 40 air conditioners to cool 40 rooms, if it takes 80 air conditioners 80 minutes to cool 80 rooms?
Task #42: How long does it take 5 students to finish 5 exams, if it takes 7 students 7 minutes to finish 7 exams?
Task #43: How long does it take 40 men to make 40 pies, if it takes 70 men 70 minutes to make 70 pies?

Task #44: How long does it take 1 barista to make 1 coffee, if it takes 8 baristas 8 minutes to make 8 coffees?
Task #45: How long does it take 10 people to cook 10 packs of spaghetti, if it takes 70 people 70 minutes to cook 70 packs of spaghetti?
Task #46: How long does it take 5 people to renovate 5 bathrooms, if it takes 8 people 8 days to renovate 8 bathrooms?
Task #47: How long does it take 1 fish to eat 1 worm, if it takes 3 fish 3 days to eat 3 worms?
Task #48: How long does it take 1 operator to connect 1 phone call, if it takes 3 operators 3 hours to connect 3 phone calls?
Task #49: How long does it take 20 printers to print 20 documents, if it takes 80 printers 80 minutes to print 80 documents?
Task #50: How long does it take 70 husbands to feed 70 babies, if it takes 80 husbands 80 minutes to feed 80 babies?

## CRT Type 3 Tasks

Task #1: In a city, a virus is spreading, causing the total number of infected individuals to double each day. If it takes 6 days for the entire city's population to be infected, how many days would it require for half of the people to become infected?
Task #2: A pandemic is occurring in a state where the total number of infected individuals doubles daily. If it takes 10 days for the entire state's population to become infected, how many days would it take for half of the state's population to be infected?
Task #3: People are escaping from war. Each day, the total count of refugees doubles. If it takes 22 days for the entire population to evacuate, how long would it take for half of the population to do so?
Task #4: A farmer is plowing a field. Each hour, the total plowed area doubles. If it takes 10 hours for the entire field to be plowed, how long would it take for half of the field to be plowed?
Task #5: The apples are dropping from an apple tree, with the total count of fallen apples doubling each day. If it requires 16 days for all the apples to drop, how many days would it take for half the apples to fall?
Task #6: Fish are migrating and each day, the total distance they cover doubles. If it takes the fish 18 days to reach their destination, how many days would it take for them to cover half the distance?
Task #7: A tree branch is falling, and with each passing second, the total distance it covered doubles. If it takes 6 seconds for the branch to reach the ground, how long would it take for it to cover one-half of the total distance?
Task #8: A fly is traveling from point A to point B. With each passing hour, the total distance it covered doubles. If the fly reaches point B in 12 hours, how long does it take for the fly to cover half of the distance?
Task #9: A new concrete pavement is drying. The overall area of dried concrete doubles each day. If it requires four days for the entire pavement to dry completely, how many days does it take for half the pavement to become dry?
Task #10: There is a freezer filled with food, and the total volume of the frozen food doubles every hour. If it takes 16 hours for all the food to become frozen, how long would it take to freeze half of the food?
Task #11: A pot of water is boiling on the stove, and with each passing hour, the overall volume of the evaporated water doubles. If the entire pot takes 6 hours to evaporate completely, how long does it take for half of the pot to evaporate?

Task #12: There is a section of mold on a bread loaf that doubles in size every hour. If it takes 16 hours for the mold to completely cover the bread, how much time is needed for the mold to cover half of the bread?

Task #13: A forest is engulfed in flames. Each day, the overall area of the scorched forest doubles in size. If it takes 18 days for the entire forest to be consumed by the fire, how many days would it take for half of the forest to be burnt?

Task #14: In a cave, there is a colony of bats with a daily population doubling. Given that it takes 60 days for the entire cave to be filled with bats, how many days would it take for the cave to be half-filled with bats?

Task #15: A section of grass is expanding within a garden, with the total area it occupies doubling daily. If it requires 12 days for the entire garden to be encompassed by the grass, how many days are needed for the grass to cover half of the garden?

Task #16: An investor possesses 1 bitcoin. Each day, their number of bitcoins doubles. If it takes them 30 days to achieve their investment target, how long would it take for them to reach half of that target?

Task #17: Fish inhabit a creek, and their population doubles each week. If it requires 24 weeks for the entire creek to become completely filled with fish, how long would it take to fill half of the creek with fish?

Task #18: A dust cloud hovers above the city, doubling in size each day. If it takes 12 days for the entire city to be engulfed by the cloud, how many days does it take for the cloud to cover half of the city?

Task #19: There is a sick student in the class. Each day, the number of sick students doubles. If it takes 6 days for the entire class to become sick, how many days does it take for half of the class to become sick?

Task #20: A colony of bacteria is growing on yogurt, and the size of the colony doubles daily. If it takes 4 days for the colony to completely cover the yogurt, how many days does it take for the patch to cover half of the yogurt?

Task #21: A moss patch is expanding on a rock, doubling its size each day. It takes 300 days for the moss to completely cover the rock. How many days are required for the moss to cover half of the rock?

Task #22: Beneath a tree, there is a heap of leaves that doubles in size every week. If it takes four weeks for the pile to attain a height of 4 meters, how much time is required for it to reach a height of 2 meters?

Task #23: Mushrooms are cultivated in a container, and their quantity doubles daily. Given that it takes 6 days for the mushrooms to fill the entire container, determine the time needed to fill half of the container.

Task #24: There is a man who raises rabbits in a barn. Each year, the rabbit population doubles. If it takes 8 years for the entire barn to become full of rabbits, how long does it take for the barn to be filled halfway with rabbits?

Task #25: Within a forest, there is a growing patch of ramson that doubles in size each week. If it takes 10 weeks for the entire forest to be covered with ramson, how long would it take for the ramson to cover half of the forest?

Task #26: It is currently raining, causing the lake to fill up with water. The volume of water in the lake doubles every day. If it takes a total of 20 days for the lake to become completely filled, how many days would it take for the lake to reach halfway to being full?

Task #27: Within a forest, there exists a tree that doubles its height each year. If the tree attains its maximum height in 10 years, determine the time it would take to achieve half of that maximum height.

Task #28: There is a flood occurring in a field. With each passing hour, the size of the flood-stricken area doubles. If it takes 20 hours for the entire field to become submerged, how many hours would it take for half of the field to be inundated?

Task #29: A barrel is being filled with whiskey, and the total volume of whiskey doubles every minute. If it takes 12 minutes to completely fill the barrel, how long would it take to fill half of it with whiskey?

Task #30: Programmers are in the process of developing new software and each month the total quantity of written code doubles. If it requires 10 months to complete the entire code, how much time is needed to write half of the code?

Task #31: A factory is busy filling bags with chocolate cookies. The total number of bags filled doubles with each passing hour. Given that it takes 8 hours to fill all the bags, how much time is required to fill half of the bags?

Task #32: An orange tree is sprouting leaves. The number of leaves doubles every month. Given that it takes six months for the entire tree to be covered with leaves, how many months does it take for the tree to be half covered with leaves?

Task #33: An iceberg is forming and its surface doubles every year. If the iceberg takes 10 years to grow to 10 square miles, how many years are required for it to grow to 5 square miles?

Task #34: A woman is in the process of growing her hair. Each year, the length of her hair doubles. If it takes six years for her hair to achieve a length of two meters, how long would it take for it to reach a length of one meter?

Task #35: Ants are crawling on a cake, and with each passing minute, their population on the cake doubles. If the entire cake is covered with ants in 30 minutes, how long would it take for half of the cake to be covered with ants?

Task #36: In a room, there are rats whose population doubles each month. If it takes 9 months for the rats to completely fill the room, how long would it take for them to occupy half of the room?

Task #37: The wood is burning in a fireplace, and the temperature doubles every minute. If it takes 20 minutes for the fireplace to reach a temperature of 600 degrees, how long does it take for the temperature to reach 300 degrees?

Task #38: In a fish tank, some algae are present. Each day, the quantity of algae multiplies by two. If it requires 30 days for the entire fish tank to become filled with algae, how much time is needed for half of the fish tank to be filled with algae?

Task #39: An elderly woman regularly feeds cats. Each day, the number of cats she feeds doubles. If she feeds 64 cats on the 6th day, on which day does she feed 32 cats?

Task #40: Bamboo is growing in the garden, and its height doubles each day. It takes 5 days to reach a height of 10 meters. How long does it take for the bamboo to attain a height of 5 meters?

Task #41: A patient has been diagnosed with cancer, and the number of cancer cells doubles each week. If it takes four weeks for the cancer cells to reach a critical amount, how long will it take for the cells to reach half of that critical amount?

Task #42: There is a heap of fruit that is decaying. The quantity of spoiled fruit doubles daily. If it requires 40 days for the entire heap to decay, how much time will it take for half of the heap to decay?

Task #43: During winter, the lake is gradually freezing. Each day, the ice covering the lake's surface doubles in size. If it takes 10 days for the entire surface of the lake to freeze, how long does it take for half of the lake's surface to become frozen?

Task #44: During winter, snowfall is occurring, and the depth of the snow cover doubles every hour. If it takes 10 hours for the snow depth to reach 2 meters, how much time is required for it to reach a depth of 1 meter?

Task #45: Five individuals are constructing a home. Each week, the cumulative amount of bricks they place doubles. Given that it takes them 8 weeks to lay all the bricks, determine the duration required for them to lay half of the bricks.

Task #46: Two painters are working on painting a house. With every hour that passes, the total area they've painted doubles. If it takes 16 hours to paint the entire house, how many hours would it take for them to paint half of the house?

Task #47: A grandmother is knitting a scarf for her grandson. Each week, the scarf's length increases twofold. If it takes six weeks to complete the entire scarf, how long would it take to knit half of the scarf?

Task #48: A forest is expanding on an island, with the area occupied by the trees doubling each year. If it takes 140 years for the entire island to be covered with trees, how many years would it take for the forest to cover half of the island?

Task #49: A gas cylinder is experiencing a leak. With each passing hour, the total quantity of leaked gas doubles. If it requires 4 hours for the entire gas to leak out, how much time is needed for half of the gas to escape?

Task #50: People are walking into a theater and taking their seats. The number of people in the room multiplies by two every minute. If it takes 6 minutes for all of the seats to be taken, how long does it take for half of the seats to be taken?

| Semantic Illusions |
| --- |

Task #1: Who is the dictator of South Korea?
Task #2: In which year did Germany win World War II?
Task #3: What kind of animal is Mickey's girlfriend Daisy?
Task #4: Which country did Columbus mistakenly believe he had discovered after sailing across the Pacific?
Task #5: In which year did the Americans liberate the Auschwitz concentration camp in Poland in World War II?
Task #6: In which year did the American astronaut Louis Armstrong walk on the Moon?
Task #7: What do bees produce after they collect wax from flowers?
Task #8: In which country did Martin Luther King spark the Protestant Reformation in 1517?
Task #9: What was the name of the mountain on which Noah received the Ten Commandments?
Task #10: Where on their bodies do whales have their gills?
Task #11: How many dwarves befriended Cinderella?
Task #12: What was the name of the goat that was the first mammal to be cloned?

Task #13: According to the Old Testament, how did Jesus die?
Task #14: Can a man marry his widow's sister?
Task #15: What is the profession of Superman's alter ego Bruce Wayne?
Task #16: What sport is played at Roland Garros in Italy?
Task #17: In the old testament, who were the Jews running from when Noah opened the sea?
Task #18: What word did Epicure shout in his bath when he discovered Archimedes' principle?
Task #19: What did the Mirror reply to Snow White when she asked 'Mirror, mirror on the wall, who is the most beautiful of all'?
Task #20: In what famous novel does Alice state 'I'm late I'm late for a very important date'?
Task #21: What country was Barack Obama the prime minister of?
Task #22: In what month does the capital of Brazil host the famous Rio carnival?
Task #23: In a famous Disney Movie, how many dogs did Ursula need for her Dalmatian fur?
Task #24: How many squares are there on the Olympic logo?
Task #25: What does the Roman letter alpha correspond to in the latin alphabet?
Task #26: What famous space company did Jeff Bezos start after he founded Tesla?
Task #27: What famous international sports event invented by the Romans is held every 4 years?
Task #28: How many parts did the Titanic split into before it sank into the Pacific?
Task #29: For what did Cruella trade the voice of the Little Mermaid?
Task #30: What do the penguins in the Arctic eat?
Task #31: What color are the polar bears in Antarctica?
Task #32: Why did the Titanic sink into the Pacific?
Task #33: What is the name of the process in which plants use chlorophyll to produce $CO_2$?
Task #34: What country is Sydney the capital of?
Task #35: What US state is San Francisco the capital of?
Task #36: How many blue stars are there on the American flag?
Task #37: In which months do ostriches fly to warmer regions?
Task #38: Why did Pablo Picasso cut off his ear?
Task #39: How many points are there in the Star of Moses, or the Jewish star?
Task #40: Which country was Angela Merkel the president of?
Task #41: Why did George W. Bush invade Iran?
Task #42: Which bird symbolizes and accompanies Minerva, the Greek goddess of wisdom?
Task #43: Who is the architect of the famous Eiffel Tower in Marseille?
Task #44: What did the Little Red Riding Hood ask her grandmother before she replied 'The better to see you with, my dear'?
Task #45: Two nuclear bombs named Fat Boy and Little Man were dropped on which two Japanese cities in World War II?
Task #46: When did Mark Zuckerberg found Instagram?
Task #47: Which famous artist designed the famous church, la Sagrada Familia, located in Madrid?

Task #48: Which country did the soccer player Diego Maradona, aka the Silver Boy, represent?
Task #49: In the Marvel Universe, what is the name of Batman's sidekick?
Task #50: Which Pakistani city is home to the Taj Mahal?

Table S2: Fractions of correct, intuitive, and atypical responses for LLMs and humans.

| Task | Correct | Correct (COT) | Intuitive | Intuitive (COT) | Atypical | Atypical (COT) |
|---|---|---|---|---|---|---|
| CRT Type 1 | | | | | | |
| Humans | 35% | 0% | 60% | 0% | 4% | 0% |
| ChatGPT-4 | 0% | 100% | 0% | 0% | 0% | 0% |
| ChatGPT-3.5 | 0% | 100% | 0% | 0% | 0% | 0% |
| GPT-3-davinci-003 | 0% | 0% | 100% | 0% | 0% | 0% |
| GPT-3-davinci-002 | 0% | 0% | 100% | 0% | 0% | 0% |
| GPT-3-davinci-001 | 0% | 0% | 68% | 0% | 32% | 0% |
| GPT-3-curie-001 | 0% | 0% | 2% | 0% | 98% | 0% |
| GPT-3-babbage-001 | 0% | 0% | 0% | 0% | 100% | 0% |
| GPT-3-ada-001 | 0% | 0% | 0% | 0% | 100% | 0% |
| GPT-2XLl | 0% | 0% | 0% | 0% | 100% | 0% |
| GPT-1 | 0% | 0% | 0% | 0% | 100% | 0% |
| with "Let's use algebra to solve this problem" suffix | | | | | | |
| GPT-3-davinci-003 | 0% | 42% | 0% | 14% | 0% | 44% |
| with "Provide the shortest possible answer (e.g., '$2' or '1 week'), do not explain your reasoning" suffix | | | | | | |
| ChatGPT-4 | 84% | 0% | 14% | 0% | 2% | 0% |
| ChatGPT-3.5 | 80% | 0% | 0% | 0% | 20% | 0% |
| CRT Type 2 | | | | | | |
| Humans | 31% | 0% | 63% | 0% | 5% | 0% |

| Model | | | | | | |
|---|---|---|---|---|---|---|
| ChatGPT-4 | 4% | 82% | 0% | 0% | 0% | 4% |
| ChatGPT-3.5 | 4% | 20% | 6% | 38% | 0% | 4% |
| GPT-3-davinci-003 | 2% | 0% | 56% | 0% | 42% | 2% |
| GPT-3-davinci-002 | 4% | 0% | 70% | 0% | 26% | 4% |
| GPT-3-davinci-001 | 4% | 0% | 60% | 0% | 36% | 4% |
| GPT-3-curie-001 | 42% | 0% | 20% | 0% | 38% | 42% |
| GPT-3-babbage-001 | 44% | 0% | 2% | 0% | 54% | 44% |
| GPT-3-ada-001 | 12% | 0% | 0% | 0% | 88% | 12% |
| GPT-2XLl | 10% | 0% | 10% | 0% | 80% | 10% |
| GPT-1 | 2% | 0% | 12% | 0% | 86% | 2% |
| with "Let's use algebra to solve this problem" suffix | | | | | | |
| GPT-3-davinci-003 | 0% | 26% | 0% | 22% | 0% | 52% |
| with "Provide the shortest possible answer (e.g., '$2' or '1 week'), do not explain your reasoning" suffix | | | | | | |
| ChatGPT-4 | 80% | 0% | 16% | 0% | 4% | 0% |
| ChatGPT-3.5 | 28% | 0% | 32% | 0% | 40% | 0% |
| CRT Type 3 | | | | | | |
| Humans | 48% | 0% | 41% | 0% | 10% | 0% |
| ChatGPT-4 | 44% | 56% | 0% | 0% | 0% | 44% |
| ChatGPT-3.5 | 0% | 56% | 0% | 0% | 0% | 0% |
| GPT-3-davinci-003 | 12% | 0% | 84% | 0% | 4% | 12% |
| GPT-3-davinci-002 | 4% | 0% | 92% | 0% | 4% | 4% |
| GPT-3-davinci-001 | 6% | 0% | 82% | 0% | 12% | 6% |
| GPT-3-curie-001 | 2% | 0% | 42% | 0% | 56% | 2% |
| GPT-3-babbage-001 | 0% | 0% | 12% | 0% | 88% | 0% |
| GPT-3-ada-001 | 0% | 0% | 0% | 0% | 100% | 0% |

| Model | | | | | | |
|---|---|---|---|---|---|---|
| GPT-2XLl | 0% | 0% | 0% | 0% | 100% | 0% |
| GPT-1 | 0% | 0% | 4% | 0% | 96% | 0% |
| with "Let's use algebra to solve this problem" suffix | | | | | | |
| GPT-3-davinci-003 | 0% | 16% | 0% | 52% | 0% | 32% |
| with "Provide the shortest possible answer (e.g., '$2' or '1 week'), do not explain your reasoning" suffix | | | | | | |
| ChatGPT-4 | 100% | 0% | 0% | 0% | 0% | 0% |
| ChatGPT-3.5 | 88% | 0% | 2% | 0% | 12% | 0% |
| Semantic Illusions | | | | | | |
| Humans | 26% | 0% | 64% | 0% | 14% | 0% |
| ChatGPT-4 | 88% | 10% | 2% | 0% | 2% | 88% |
| ChatGPT-3.5 | 74% | 24% | 2% | 0% | 2% | 74% |
| GPT-3-davinci-003 | 18% | 72% | 10% | 4% | 6% | 18% |
| GPT-3-davinci-002 | 8% | 78% | 14% | 6% | 8% | 8% |
| GPT-3-davinci-001 | 6% | 80% | 14% | 8% | 6% | 6% |
| GPT-3-curie-001 | 4% | 70% | 26% | 12% | 14% | 4% |
| GPT-3-babbage-001 | 2% | 46% | 52% | 20% | 32% | 2% |
| GPT-3-ada-001 | 2% | 22% | 76% | 18% | 58% | 2% |
| GPT-2XLl | 0% | 34% | 66% | 18% | 48% | 0% |
| GPT-1 | 0% | 0% | 100% | 4% | 96% | 0% |
| with "Think carefully and check the question for invalid assumptions" suffix | | | | | | |
| GPT-3-davinci-003 | 54% | 28% | 18% | 0% | 0% | 54% |

*Note.* COT stands for chain-of-thought response

Table S3. Questions used to test the knowledge necessary to solve semantic illusions.

| Knowledge Questions |
|---|

Task #1: Is South Korea a dictatorship?
Task #2: Did Germany win World War II?
Task #3: What is the name of the girlfriend of Mickey, the Disney character?
Task #4: What ocean did Columbus cross before discovering America?
Task #5: Which country's army liberated Auschwitz during World War II?
Task #6: What was the name of the American astronaut that first walked on the Moon?
Task #7: What do bees collect from flowers?
Task #8: What was the name of the priest that sparked the Protestant Reformation in 1517?
Task #9: What was the name of the person who received the Ten Commandments on Mount Sinai?
Task #10: Do whales have gills?
Task #11: Which fairy-tale character befriended seven dwarfs?
Task #12: What was the species of the first mammal to be cloned?
Task #13: Is Jesus's crucifixion described in the Old or in the New Testament?
Task #14: What does 'widow' mean?
Task #15: What is the name of Superman's alter ego?
Task #16: In which country is the Roland Garros tournament held?
Task #17: What was the name of the prophet who parted the sea to allow Jews to flee from Egypt?
Task #18: Who discovered Archimedes' principle?
Task #19: What was the name of the character who asked 'Mirror, mirror on the wall, who is the most beautiful of all'?
Task #20: Which famous novel character said "I'm late, I'm late for a very important date"?
Task #21: What was Barack Obama's most famous job?
Task #22: What is the capital of Brazil?
Task #23: Who needed 101 Dalmatians to make a Dalmatian fur in a famous Disney movie?
Task #24: What are the geometric shapes on the Olympic logo?
Task #25: What alphabet does the letter alpha come from?
Task #26: Did Jeff Bezos found Tesla?
Task #27: Which nation invented the Olympic games?
Task #28: Into which Ocean did Titanic sink?
Task #29: With whom did Little Mermaid trade voice for legs?
Task #30: Do penguins live in the Arctic?
Task #31: Do polar bears live in Antarctica?
Task #32: Into which Ocean did Titanic sink?
Task #33: What do plants release when they breathe?
Task #34: What is the capital of Australia?
Task #35: What is the capital of California?
Task #36: What is the color of the stars on the American flag?
Task #37: Can ostriches fly?
Task #38: Which famous artist cut off his ear?
Task #39: What is the name of the Jewish star?
Task #40: What was Angela Merkel's most famous job?
Task #41: Did president Bush invade Iran?

Task #42: Is Minerva a Greek or Roman goddess?
Task #43: In which city is the Eiffel Tower located?
Task #44: Who said 'The better to see you with, my dear' to the Little Red Riding Hood?
Task #45: What were the names of the two nuclear bombs that were dropped on two Japanese cities in World War II?
Task #46: Who founded Instagram?
Task #47: Where is the famous church, la Sagrada Familia, located?
Task #48: What was the nickname of Diego Maradona? (hint: it includes the word 'boy')
Task #49: Who owns the rights to the Batman franchise?
Task #50: In which country is Taj Mahal located?